# A Multi-Phase Analysis of Blood Culture Stewardship: Machine Learning Prediction, Expert Recommendation Assessment, and LLM Automation

Authors: Fatemeh Amrollahi*[1],PhD, Nicholas Marshall*[2],MD, Fateme Nateghi Haredasht[1],PhD, Kameron C Black[2], MD, Aydin Zahedivash[2],MD, Manoj V Maddali[1,2],MD,PhD, Stephen P. Ma[2],MD,PhD, Amy Chang[2],MD PharmD, Stanley C Deresinski[2],MD, Mary Kane Goldstein[3],MD, Steven M. Asch[2],MD, Niaz Banaei[2,4], MD, Jonathan H Chen[1,2,**],MD,PhD

1.Stanford Center for Biomedical Informatics Research, Stanford University, Stanford, CA, USA; 2.Department of Medicine, Stanford University School of Medicine, Stanford, CA, USA;3. Department of Health Policy, Stanford University School of Medicine, Stanford, CA, USA;4.Department of Pathology, Stanford University School of Medicine, Stanford, CA, USA

## I. Abstract:

Blood cultures are often overordered without clear justification, straining healthcare resources and contributing to inappropriate antibiotic use—pressures worsened by the global shortage. In study of 135,483 emergency department (ED) blood culture orders, we developed machine learning (ML) models to predict the risk of bacteremia using structured electronic health record (EHR) data and provider notes via a large language model (LLM). The structured model's AUC improved from 0.76 to 0.79 with note embeddings and reached 0.81 with added diagnosis codes. Compared to an expert recommendation framework applied by human reviewers and an LLM-based pipeline, our ML approach offered higher specificity without compromising sensitivity. The recommendation framework achieved sensitivity 86%, specificity 57%, while the LLM maintained high sensitivity (96%) but overclassified negatives, reducing specificity (16%). These findings demonstrate that ML models integrating structured and unstructured data can outperform consensus recommendations, enhancing diagnostic stewardship beyond existing standards of care.

## II. Introduction

Blood cultures are the gold standard for diagnosing bacteremia, detecting a wide range of pathogens and providing antimicrobial susceptibility results that directly guide treatment. They remain the only universally endorsed diagnostic tool capable of identifying diverse pathogens while informing targeted treatment. Guidelines from the Infectious Diseases Society of America (IDSA) and American Society for Microbiology (ASM) emphasize the importance of collecting blood cultures promptly, ideally before antimicrobial initiation, to maximize diagnostic yield[1]. Evidence shows that pre-antibiotic cultures are significantly more likely to detect pathogens, whereas delays reduce yield and may lead to prolonged empiric therapy[2].

Although rapid molecular diagnostics are emerging, their restricted pathogen detection (limited to predefined targets), absence of antimicrobial susceptibility information, high cost, and challenges in distinguishing viable pathogens from non-viable microbial DNA or host-ingested organisms limit their use to adjunctive roles [3,4]. At present, no viable substitute exists for conventional blood cultures.

The recent global shortage of blood culture bottles placed unprecedented strain on diagnostic capacity, forcing institutions to implement rationing strategies such as limiting culture sets, imposing hard stops on repeat testing, or focusing testing efforts on patients more likely to benefit, based on clinical assessment [5,6,7]. These strategies, while necessary, introduce clinical risk. Despite the critical role of blood cultures, fewer than 10% yield true positives, and an estimated 60% are obtained without strong clinical indication [6]. Excessive or low-yield testing increases contamination rates, drives unnecessary antibiotic use, and adds burden to already strained laboratory and clinical resources [8,9].

* Equally contributed

** Corresponding author

Optimizing blood culture utilization requires balancing the need to detect serious infections against the harms of overuse and the constraints of limited supply. Ideally, the decision to obtain a blood culture should be guided by pre-test probability, integrating clinical presentation and presumptive diagnosis [9]. However, in high-pressure environments like emergency departments (ED), these decisions are often made rapidly and with incomplete information. Expert recommendations exist to guide culture collection based on presumptive diagnosis, but these frameworks are difficult to operationalize prospectively and lack the precision needed for individualized decision-making [6,7,9]. Furthermore, they are typically derived from small samples or expert opinion, limiting their scalability and reliability in diverse real-world settings.

Recent advances in the availability of electronic health record (EHR) data and machine learning (ML) methods offer an opportunity to modernize this decision process. By training predictive models on large, real-world datasets, we can develop data-driven rules that augment existing practices with more personalized decision support. These tools have the potential to support real-time clinical decision-making, improve diagnostic precision, and advance antimicrobial stewardship — particularly in the context of constrained diagnostic resources. There is a critical need for models that are both accurate and implementable, capable of identifying patients most likely to benefit from blood culture testing while minimizing unnecessary use. Our study introduces a novel, data-driven approach to optimize blood culture decision-making by leveraging structured EHR data, unstructured provider notes, and machine learning techniques. Unlike previous studies that rely solely on expert-driven heuristics or small-scale analyses, our work systematically evaluates both ML-based predictions and expert recommendations.

### III. Study Cohort

We conducted a retrospective analysis of EHR data from 135,483 ED blood culture orders, with the primary outcome being a positive result. This study was approved by the Institutional Review Board (IRB) of Stanford university. IRB information will be provided if the paper is accepted. The study included patients aged 18 years or older who had blood culture collected during their ED visit, provided they had no positive blood cultures within the preceding 14 days. Cultures that were marked with errors, discontinued, or canceled were excluded from the analysis. The primary outcome of interest was a positive blood culture result, excluding likely contaminants based on national guidelines and local microbiology protocol, including coagulase-negative staphylococci, diphtheroids, and *Bacillus* spp...

For this study, to ensure broad applicability across various healthcare systems, we utilized primary laboratory andclinical variables commonly measured in care settings. Categorical features were one-hot encoded to ensure compatibility with the predictive model. The data was split chronologically into training (orders taken between 2015-2022), development (orders taken between 2022-2023), and evaluation sets (orders taken 2023 onwards), allowing the most recent data to be used for evaluation and providing a more accurate representation of system performance in real-world settings. Data used for this dataset are available by Stanford medicine Research data Repository (STARR) [10], and codes are available at [https://github.com/HealthRex/CDSS/tree/master/scripts/Blood_Culture_Stewardship].

### IV. Model Development

The predictive model was implemented using XGBoost, with class weighting applied to address the imbalance in positive blood culture cases. A grid search on the validation set was used to fine-tune the hyperparameters. The best performance was achieved with a maximum tree depth of 4 and 30 boosting iterations. This approach ensured the model remained interpretable while maintaining high sensitivity for identifying high-risk cases, even with constrained resources.

### V. Experiment 1: Integrating Clinical Notes into Blood Culture Prediction Models

Incorporating clinical notes alongside traditional EHR variables further enhanced the model's ability to deliver real-time, evidence-based decision support in emergency department settings. Of the 135,000 patients with blood cultures, 130,983 had accompanying ED provider notes. To extract latent information from clinical notes, we utilized the STELLA 1.5 billion parameter pre-trained language model. After cleaning and filtering the notes, batches of notes were processed sequentially. Each note was tokenized with truncation at 2048 tokens to manage input length. The tokenized inputs were then passed through the model, and embeddings with the size of 1536 were derived by averaging the hidden states of the transformer outputs across the tokenized sequence, weighted by the attention mask.

To further assess the value of adding latent information to our model, we developed four predictive models to evaluate the impact of different feature sets on performance using Roc-AUC. We further report the specificity at two different sensitivity level (see Figure 1).

• Structured Model: Uses only structured EHR data including vital signs, laboratory values, and demographic information.

• BactoRisk: Expands the structured model by incorporating note embeddings extracted from ED provider notes.

ICD codes are not always accurate or consistently available, but to have a comprehensive review of all models, we also ran experiments for these two models:

• Structured + Diagnosis Code Model: Builds on the structured model by adding ICD9 and ICD 10 diagnosis codes that were active or diagnosed before culture order.

- BactoRisk + Diagnosis Code Model: Builds on the BactoRisk model by adding ICD9 and ICD 10 diagnosis codes that were active or diagnosed before culture order.

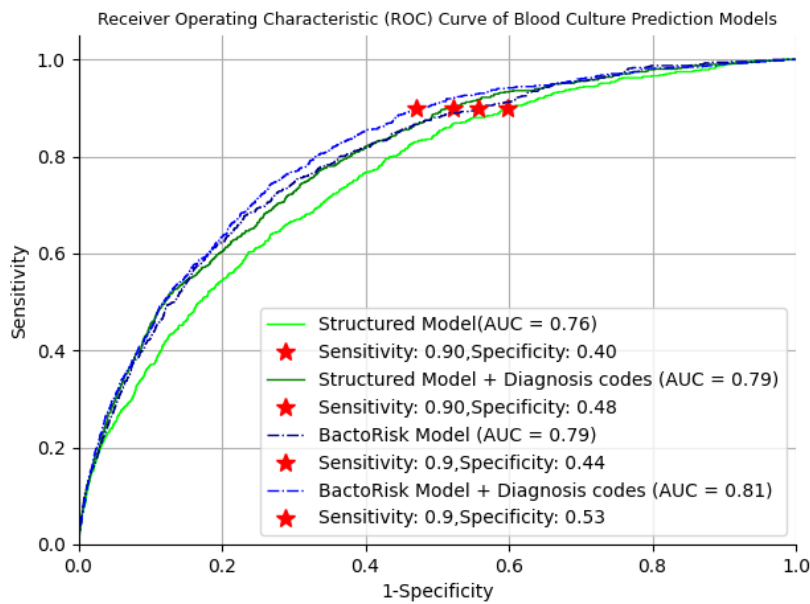

**Figure 1. Comparison of Predictive Model Performance for Bacteremia Risk Stratification.** This figure presents the ROC-AUC curves for four predictive models evaluating the impact of different feature sets. Sensitivity is marked at 90% to illustrate variations in specificity across models.

## VI. Experiment 2: Manual Risk Stratification Using Expert Recommendation Framework

We further assessed the expert recommendation framework described by Fabre et al. to identify patients at high risk for bacteremia, focusing exclusively on initial blood culture (BCx) orders [9]. This framework categorizes conditions into low, intermediate, and high risk for bacteremia, in an effort to guide blood collection when clinically indicated.

**Review Process**

To evaluate the performance of the blood culture expert recommendation framework, we engaged four expert clinicians to review 109 cases. The reviewers were divided into two groups: Dr. N. Marshall and Dr. M. Maddali assessed the first 60 cases, while Dr. K. Black and Dr. A. Zahedivash reviewed the remaining 49 cases. In cases of disagreement, Dr. J. Chen served as the tie-breaker for the first round of reviews, and Dr. N. Marshall acted as the tie-breaker for the second round.

Cases were randomly selected from patients who had blood cultures ordered in ED from 2023 onward. Both positive and negative cases were randomly sampled, with 80% of the cases being positive, as these are more challenging to assess.

*Performance of the Blood Culture Expert Recommendation Framework Using Human Judges*

According to the blood culture expert recommendation framework, blood cultures are primarily advised for patients classified as intermediate or high risk. Based on our analysis, this approach would have resulted in missing 14 patients with positive blood cultures out of 100 cases while also leading to three unnecessary tests for patients with negative blood culture results. Consequently, our findings indicate that the overall accuracy, sensitivity, specificity, and F1 score of the guideline is 74%, 86%, 57% and 79%, respectively.

The stratification of the reviewed cases is provided in **Table 1**

Table 1. Case stratification by human judges using the blood culture expert recommendation framework

|  | Patients with positive blood culture Results | | | | Unclassified |
|---|---|---|---|---|---|
|  | Total | Low Risk | Intermediate Risk | High Risk |  |
| N patients | 100 | 14 | 37 | 50 | 0 |
|  | Patients with Negative blood culture Results | | | | |
| N patients | 9 | 4 | 1 | 2 | 2 |

*Performance of the Blood Culture Expert Recommendation Framework Using a Large Language Model for Patient Evaluation*

To further assess the performance of the blood culture expert recommendation framework, we implemented a pipeline leveraging a HIPAA-compliant GPT-4 model[11] to evaluate bacteremia risk. This approach automates risk stratification by integrating the guideline into a structured prompt-based framework. For each case, we provided the model with blood culture expert recommendation framework, patient EHR data, and ED provider notes recorded at the time of blood culture ordering. The full prompt structure is illustrated in Figure 2.

By incorporating this evidence-based framework, our model aligns predictive outputs with established clinical recommendations, enhancing accuracy, reducing false-positive cultures, and optimizing antimicrobial stewardship. The primary unit of analysis in this experiment was the **patient-encounter-order**, representing each unique culture order linked to a patient encounter within the ED. Given the cost associated with each API call, we randomly selected 1,000 patient-encounter-order instances for evaluation. While the prevalence of positive blood cultures in our study cohort was 19%, we intentionally oversampled positive cases to improve model assessment. For cases where the pipeline produced invalid outputs, we repeated the assessment to ensure result validity.

As shown in Table 2, our model correctly identified 393 out of 409 cases at high risk for bacteremia. However, 16 high-risk cases were missed. Additionally, our findings indicate a tendency for the model to over-classify patients as high risk, as 315 out of 591 patients with negative blood cultures were incorrectly classified as high risk. Overall, the guideline-based approach achieved an **accuracy of 56%**, **sensitivity of 96%**, **specificity of 16%**, and an **F1 score of 68%**.

While both human reviewers and the LLM-based approach assessed bacteremia risk using the blood culture expert recommendation framework, their performance demonstrated notable differences. Human judges yielded a higher overall accuracy (74% vs. 56%) and specificity (57% vs. 16%), suggesting better discernment in avoiding unnecessary blood cultures. However, the LLM exhibited superior sensitivity (96% vs. 86%), indicating a stronger tendency to identify high-risk cases and minimize missed bacteremia diagnoses. Despite its high sensitivity, the LLM over-classified patients as high risk, leading to a higher false-positive rate (315 misclassified cases out of 591 negative cultures). Given this tendency, incorporating expert clinician review of LLM assessments may help refine risk stratification and reduce unnecessary testing while still leveraging the model's strength in identifying high-risk patients. In our next study we will assess the hybrid approach which could optimize clinical decision-making by balancing sensitivity and specificity more effectively.

Table 2. Performance Evaluation of the Blood Culture Expert Recommendation Framework Using a Large Language Model

| | Patients with positive blood culture Results | | | | Unclassified |
|---|---|---|---|---|---|
| | Total | Low Risk | Intermediate Risk | High Risk | |
| N Orders | 409 | 16 | 64 | 329 | - |
| | Patients with Negative blood culture Results | | | | |
| N Orders | 591 | 100 | 189 | 315 | - |

**VII. Discussion**

This study presents a multi-step analysis of blood culture decision-making using machine learning and expert recommendation-based approaches, addressing the pressing need for diagnostic stewardship, particularly during a global shortage of blood culture bottles. Leveraging a large and diverse electronic health record (EHR) dataset from over 130,000 emergency department (ED) encounters, we developed and validated predictive models and evaluated widely endorsed expert recommendations to assist blood culture collection practices.

Our first experiment demonstrated that a machine learning model built on structured EHR data alone achieved strong predictive performance (AUC = 0.76) for blood culture positivity. Incorporating nuanced clinical insights extracted from ED provider notes using a large language model (LLM) further improved performance (AUC = 0.79), particularly in specificity, without sacrificing sensitivity. These findings suggest that narrative documentation contains critical diagnostic signals (e.g., clinical reasoning, symptoms, and physical exam findings) that improve the model's ability to differentiate between positive and negative results beyond structured variables alone. Further addition of diagnosis codes increased the model's AUC to 0.81, though diagnosis codes are often unavailable or inaccurate at the time of decision-making. Collectively, these results support the development of real-time, evidence-based clinical decision support tools that incorporate both structured and unstructured EHR data to optimize blood culture use.

In our second experiment, we evaluated the performance of a commonly cited expert recommendation for blood culture collection, which stratifies clinical conditions into low, intermediate, or high risk for bacteremia and generally recommends cultures for intermediate- or high-risk scenarios. Among manually reviewed ED cases, this approach achieved 86% sensitivity and 57% specificity but would have missed 14% of patients with true positive blood cultures — primarily those classified as low risk. These findings highlight important limitations in translating expert recommendations to real-world practice, particularly in complex or atypical presentations where strict application of pre-test probability thresholds may not adequately capture diagnostic nuance. Moreover, the observed specificity suggests that a considerable number of low-yield cultures would still be collected, potentially undermining stewardship efforts.

To scale this evaluation, our third experiment applied the expert recommendation framework using a HIPAA-compliant LLM to automate risk stratification across a broader cohort. While the LLM approach preserved high sensitivity (96%), it demonstrated a marked reduction in specificity (16%), resulting in over-classification of negative cases as high risk. This over-sensitivity suggests a cautious bias in the LLM's application of expert frameworks and reflects a tendency to err on the side of overclassification in the absence of more granular clinical reasoning. Although promising for reducing missed bacteremia, the associated false-positive rate diminishes its effectiveness as a stand-alone stewardship tool.

The comparative performance of human versus LLM-based application of expert recommendations provides important insights. Human reviewers achieved higher overall accuracy and specificity, likely due to their ability to integrate subtle clinical cues, weigh uncertainty, and contextualize risk in the broader diagnostic picture. In contrast, the LLM was tuned toward high sensitivity, prioritizing safety at the expense of over-testing. These complementary strengths suggest that a hybrid approach, using LLMs to flag potentially high-risk or ambiguous cases for targeted clinician review, may optimize diagnostic precision while conserving resources.

One limitation of our study is its generalizability. To enhance the robustness of our findings, we are actively expanding our evaluation to two additional healthcare systems. This ongoing work will provide valuable insights into the model's performance across diverse clinical environments.

 The blood culture expert recommendation framework, while adopted by multiple institutions and supported by national stakeholders, has not been prospectively validated. Our model, by contrast, can be deployed in real-time settings, and we have developed a pipeline to silently test its performance in parallel with clinical workflows.

Results from this ongoing real-time validation effort will be presented in a future study. Additionally, while interrater agreement was strong, expert review introduces some degree of subjectivity, particularly in intermediate-risk cases.

### VIII.     Conclusion

This study demonstrates that machine learning models integrating both structured EHR data and unstructured clinical documentation can support more precise blood culture decision-making than widely used expert recommendation frameworks alone. Our model, which incorporates embeddings from ED provider notes using a large language model (LLM), achieved improved specificity while maintaining high sensitivity, highlighting the diagnostic value of narrative clinical data. Compared to expert-driven frameworks, which showed lower specificity and missed a proportion of true bacteremia cases, particularly in intermediate-risk patients, our findings support a shift toward individualized, data-driven support tools that better reflect real-world complexity.

While expert recommendations remain a helpful foundation, they are often limited by subjectivity, rigidity, and implementation challenges in dynamic clinical settings. In contrast, ML-enhanced tools offer scalable, context-aware decision support. The comparative performance of human and LLM-applied frameworks illustrates the trade-offs between sensitivity and specificity and suggests that a hybrid approach, where machine learning models assist with risk stratification and flag ambiguous cases for clinician oversight, may offer an optimal balance. These findings support the use of data-driven tools as a mechanism for diagnostic stewardship, improving the precision of blood culture utilization and ultimately enhancing patient care.

**Instruction:** Assume you are a physician evaluating a patient for the risk of bacteremia using the provided patient records and emergency department (ED) notes. Classify the patient into one of the following risk categories. You must only reply with one of the following options: High Risk, Intermediate Risk, Low Risk, Unclassified (if none of the categories apply)

High Risk: Patients exhibiting signs of severe sepsis or septic shock, such as hypotension, elevated lactate levels, altered mental status, or organ dysfunction (e.g., acute kidney injury, liver dysfunction). Additionally, patients with implanted prosthetic devices (e.g., orthopedic or intravascular prostheses) or conditions predisposing them to endovascular infections (e.g., immunosuppression, recent invasive procedures, known endocarditis, or vascular infections) should be classified as high risk.

Intermediate Risk: Patients presenting with symptoms suggestive of systemic infection (e.g., fever, chills, elevated white blood cell count, localized infectious process) but without overt signs of severe sepsis or septic shock. This category may include patients with localized infections (e.g., pyelonephritis, cellulitis, pneumonia, or abscesses) that have the potential to become systemic.

Low Risk: Patients with nonbacterial processes or syndromes that carry a low risk of bacteremia, especially in the absence of systemic signs of infection or risk factors for endovascular infections. Patients in this category typically lack fever, leukocytosis, and other significant signs of infection.

**ED Note:** The patient is a 42-year-old male presenting to the emergency department with a three-day history of persistent right lower abdominal pain, gradually worsening and associated with intermittent nausea and decreased appetite. He reports a low-grade fever at home but denies chills or rigors. His past medical history is significant for recurrent kidney stones and a prior appendectomy. He denies recent urinary symptoms but notes mild dysuria a week ago that resolved on its own. No recent travel or sick contacts. He has a history of hypertension managed with lisinopril but is otherwise in good health. The patient describes the pain as sharp and localized to the right lower quadrant, exacerbated by movement and partially relieved with over-the-counter analgesics. He denies vomiting, diarrhea, or significant changes in bowel habits.

**Most Recent Lab Values:** WBC: 11.3  NA: 132  BUN: 17  Cr: 0.92  Glucose: 395  UA-RBC: 0-3  UA-WBC: 51-100

**Model output: Intermediate Risk**

---

**Instruction:** Assume you are a physician evaluating a patient for the risk of bacteremia using the provided patient records and emergency department (ED) notes. Classify the patient into one of the following risk categories. You must only reply with one of the following options: High Risk, Intermediate Risk, Low Risk, Unclassified (if none of the categories apply)

High Risk: Patients exhibiting signs of severe sepsis or septic shock, such as hypotension, elevated lactate levels, altered mental status, or organ dysfunction (e.g., acute kidney injury, liver dysfunction). Additionally, patients with implanted prosthetic devices (e.g., orthopedic or intravascular prostheses) or conditions predisposing them to endovascular infections (e.g., immunosuppression, recent invasive procedures, known endocarditis, or vascular infections) should be classified as high risk.

Intermediate Risk: Patients presenting with symptoms suggestive of systemic infection (e.g., fever, chills, elevated white blood cell count, localized infectious process) but without overt signs of severe sepsis or septic shock. This category may include patients with localized infections (e.g., pyelonephritis, cellulitis, pneumonia, or abscesses) that have the potential to become systemic.

Low Risk: Patients with nonbacterial processes or syndromes that carry a low risk of bacteremia, especially in the absence of systemic signs of infection or risk factors for endovascular infections. Patients in this category typically lack fever, leukocytosis, and other significant signs of infection.

**ED Note:** The patient is a 91-year-old male with a history of hypertension, hypothyroidism, lymphoma, and chronic urinary retention, who was brought to the emergency department by his son from an assisted living facility after he accidentally dislodged his urinary catheter. He has had a catheter in place for the past six weeks and does not recall how it was removed. He denies any penile or abdominal pain but presents with a fever of 101.8°F. He has no complaints of cough, sore throat, runny nose, or abdominal pain. On examination, he appears elderly but is not in distress. His respiratory and cardiovascular exams are unremarkable, and his abdomen is soft and non-tender.

**Most Recent Lab Values:** WBC: 11.7  NA: 128  BUN: 22  Cr: 0.86  UA-RBC: >100  UA-WBC: >100

**Model output: High Risk**

**Figure 2.** (a) An example of a false positive case where the LLM misclassifies a patient with low risk of bacteremia as intermediate risk. (b) An example of our prompt and a case correctly classified as high risk for bacteremia.